\documentclass[runningheads]{llncs}
\usepackage{amsmath,amssymb,amsfonts}
\usepackage{algorithmic}
\usepackage{graphicx}
\usepackage{textcomp}
\usepackage{xcolor}
\usepackage{multicol}
\usepackage{multirow}
\usepackage{dblfloatfix} 
\usepackage{float}
\usepackage{subfig}  
\usepackage{amsmath}
\usepackage[pagebackref=true]{hyperref}
\hypersetup{
     colorlinks   = true,
     citecolor    = blue
}

\begin{document}
\title{Advanced Capsule Networks via Context Awareness}
\author{Nguyen Huu Phong\orcidID{0000-0002-5022-0226}* \and
Bernardete Ribeiro\orcidID{0000-0002-9770-7672}}
\authorrunning{Phong N.H. and Bernardete R.}
\titlerunning{Advanced Capsule Networks via Context Awareness}
\institute{CISUC, Department of Informatics Engineering, University of Coimbra, Portugal
\email{\{phong,bribeiro\}@dei.uc.pt}}
\maketitle              
\begin{abstract}
Capsule Networks (CN) offer new architectures for Deep Learning (DL) community. Though its effectiveness has been demonstrated in MNIST and smallNORB datasets, the networks still face challenges in other datasets for images with distinct contexts. In this research, we improve the design of CN (Vector version) namely we expand more Pooling layers to filter image backgrounds and increase Reconstruction layers to make better image restoration. Additionally, we perform experiments to compare accuracy and speed of CN versus DL models. In DL models, we utilize Inception V3 and DenseNet V201 for powerful computers besides NASNet, MobileNet V1 and MobileNet V2 for small and embedded devices. We evaluate our models on a fingerspelling alphabet dataset from American Sign Language (ASL). The results show that CNs perform comparably to DL models while dramatically reducing training time. We also make a demonstration and give a link for the purpose of illustration.
\keywords{Capsule Networks \and Deep Learning \and Transfer Learning \and Demonstration.}
\end{abstract}
\section{Introduction}
\label{sec:Introduction}
Capsule Networks arrive in the field of Deep Learning at the time when many issues are considered solved e.g. in image and object recognition, very deep networks with hundreds of layers are able to outperform human. One reason behind the success of DL is the Max Pooling (MP) layer which not only reduces dimension of images but also selects most critical pixels for routing from one layer to another. Though MP works very well, Hinton argues that MP causes loss of useful information. As a consequence, this layer is replaced with a routing algorithm and a new architecture namely Capsule Networks is designed~\cite{sabour2017dynamic}. The CNs can also be referred as Vector CNs since the approach is based on agreements between vectors. The design achieved $0.25\%$ test error on MNIST dataset in comparison with the state of the art using DropConnect $0.39\%$ without data augmentation~\cite{wan2013regularization}. The other CN applying Expectation Maximization (Matrix CN) reduces the best error rate by $45\%$ on SmallNORB dataset~\cite{hinton2018matrix}. However, the networks still face challenges on other datasets e.g. Cifar-10, SVHN and ImageNet.

In this research, we choose to build our CNs based on the Vector CN architecture since there are few implementations in literature for Matrix CNs and our preliminary results show that the latter takes longer convergence time in the ASL alphabet dataset.

For DL architecture, we choose several prestige Transfer Learning (TL) models including Inception V3 \cite{szegedy2016rethinking}, DenseNet V201 \cite{huang2017densely}, NASNet \cite{zoph2018learning}, MobileNet V1 \cite{howard2017mobilenets} and MobileNet V2 \cite{sandler2018mobilenetv2} to generate feature maps and compare these models to explore which model is the best in our setting. As TL is a very fast growing field, the models are trained on a wide variety of platforms such as Tensorflow, Caffe, Torch and Theano. In addition, just a year ago, Keras which was one of the biggest independent platforms had been integrated into Tensorflow. As Keras includes pre-trained models for most of leading TLs, we prefer to use this platform to set a unified environment for comparison of all TL models. We also classify models in two groups one that is mainly used for demanding computers (Inception V3, DenseNet V201) and the other for smaller devices (NASNet, MobileNet V1 and V2) since mobiles have become a crucial tool in our daily life.

We perform experiments on static signs of ASL dataset. In ASL, there are two distinct signs namely dynamic signs and static signs. Our aim is to build an action recognition framework to recognize signs in continuous frames (e.g. transcription generators for ASL songs or conversations). In this work, we focus on ASL alphabet signs. For this problem, models are usually based on Convolution Networks~\cite{ameen2017convolutional,bheda2017using}, other Machine Learning techniques like Multilayer Random Forest \cite{kuznetsova2013real} or Transfer Learning models such as GoogLeNet and AlexNet~\cite{kang2015real,garcia2016real}.

The rest of this article is structured as follows. We highlight our main contributions in Section~\ref{sec:contrib}. Next, we describe an ASL dataset for this research in Section ~\ref{sec:dataset}. Then we discuss about Vector CNs and TL models' architectures in Section~\ref{sec:capsule} and Section~\ref{sec:transferlearning}, accordingly. Experiments and respective results are analyzed and discussed in Section~\ref{sec:experiments}. We conclude this work in Section~\ref{sec:conclude}.
\section{Contribution}
\label{sec:contrib}
In our research, we improve the design of the Vector Capsule Networks and perform empirical comparisons versus Deep Learning models on accuracy and speed using an ASL fingerspelling alphabet. 

First, we propose to modify the Vector CN's architecture to find the most efficient designs. Namely, we extended Convolution layers to better filter input images and varied Fully Connected layers in the Reconstruction to leverage image restoration.

Second, we explore distinct Transfer Learning models for demanding devices (Inception V3 and DenseNet V201) and small devices (NasNet and MobileNets) when integrated with Multilayer Perceptron and Long Short Term Memory. We use the former setting as a baseline to compare with the latter.

Third, Vector CNs are analyzed against DL models. We find out that CNs perform comparatively on both accuracy and speed. In addition, CNs have an advantage as pre-training is not required. Capsules can be trained within an hour compare to days or more for pre-trained TL models.

Finally, we make a demonstration to compare CNs and DLs on videos for teaching ASL alphabets. The results are very promising as our models can recognize almost of all signs without previously seen.
\section{Recognition Models}
\label{sec:recognition}
In this section, we first discuss about an ASL dataset that will be used in our experiments. Then we deal with architectures of Vector CNs and DL models.
\subsection{ASL Dataset}
\label{sec:dataset}
\begin{figure*}[h]
\begin{center}
\includegraphics[width=0.95\textwidth]{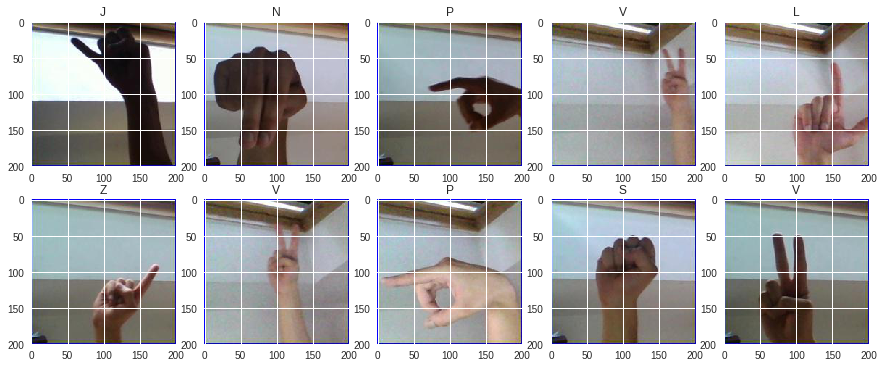}
\caption{Random Samples from ASL Dataset}
\label{fig:ASL_Dataset}
\end{center}
\end{figure*}
One of our ultimate goals is to build an ASL translator that is capable of classifying alphabet signs from language training videos. In these videos, professional trainers illustrate hand shapes for signs from A to Z. As they perform demonstrations, the hand is moving around the screen from left to right, up to down and vice versa. To train our models, we search for a dataset that captures similar movements with a large sample size. We expect that each sign should have thousands of instances since a standard MNIST has $60000$ samples for $10$ classes. With these constraints and since ASL datasets are relatively fewer than for English alphabet, we found only one dataset from Kaggle website\footnote{https://www.kaggle.com/grassknoted/asl-alphabet} that meets our requirements.
This set of data includes all 26 signs including dynamic signs ``J" and ``Z" with 3 additional signs.
Each sign includes $3000$ samples ($200 \times 200$ pixels), totally $87000$ for all signs. In these images, hands are placed in distinct positions on the screen, distance and lightning are varied.

Figure~\ref{fig:ASL_Dataset} shows 10 random samples from the dataset. We notice that signs ``N'' and ``P" have different shapes compared to the target video. In stead of replacing them, these signs are retained for the reason mentioned above.

The data are selected randomly and split into training and testing sets with the ratio of $70$/$30$. In addition, the value range is re-scaled from $[0,255]$ to $[0,1]$. For data augmentation, the rotation, shear, width shift and height shift are set in the ranges of $20$, $0.2$, $0.2$ and $0.2$ respectively. Moreover, images' brightness and contrast are spanned using random uniform within distances of $0.6$ and $1.5$.

\subsection{Capsule Networks}
\label{sec:capsule}
\begin{figure}[h]
\begin{center}
\includegraphics[width=0.95\textwidth]{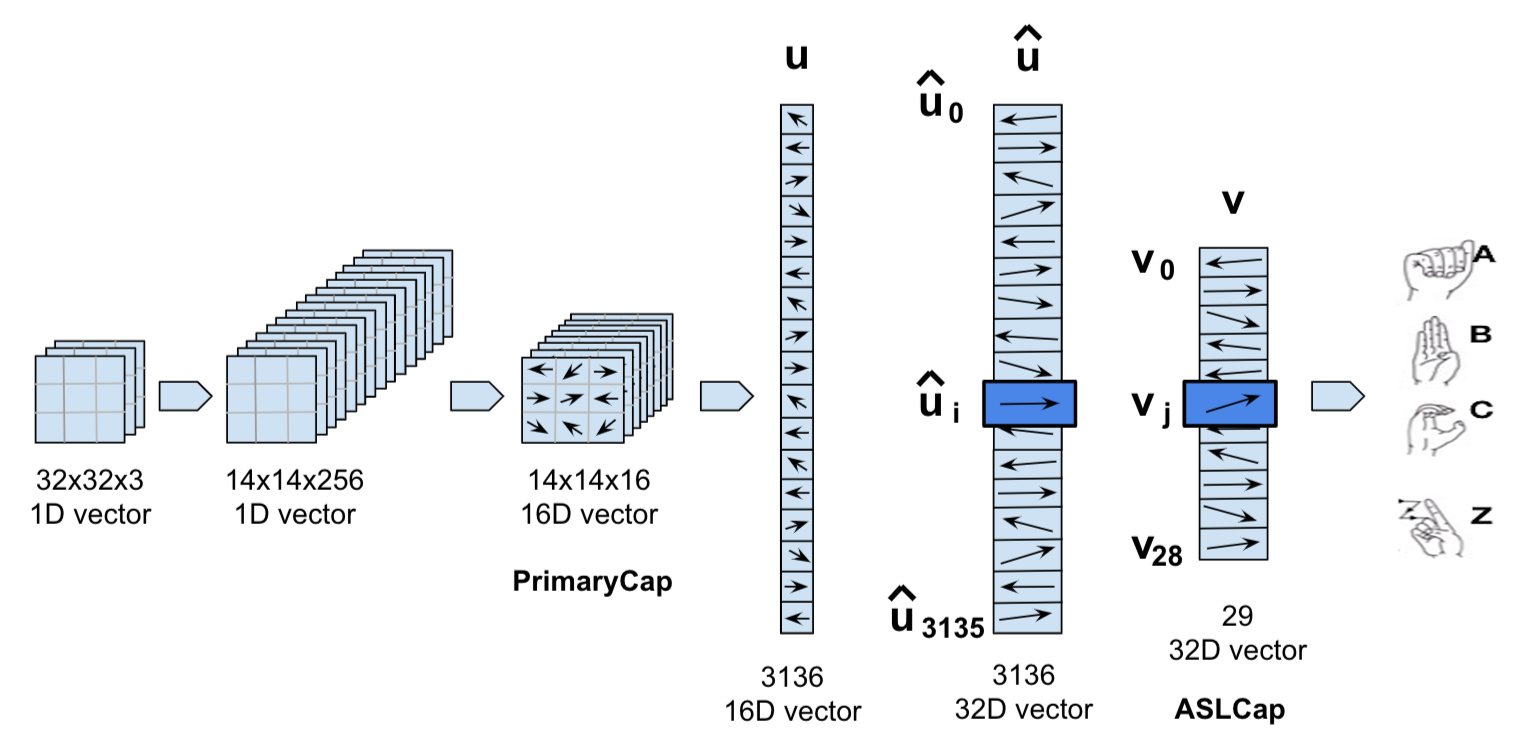}
\caption{Vector Capsule Networks Architecture for ASL}
\label{fig:asl_capsule}
\end{center}
\end{figure}
Hinton describes a Capsule Networks as a group of neurons that represent distinct properties of the same entity. In Vector CN, a capsule in one layer sends its activity to the capsule in above layer and the method checks which one agrees the most. The essential structure of Capsule Networks is shown in the Figure \ref{fig:asl_capsule}.

The initial step of Vector CNs is similar to Convolutional Neural Networks where input images are filtered to detect features such as edge and curve. Then, in PrimaryCaps, the generated features are grouped to create multi-dimension vectors. Illustrated in the Figure, $256$ feature maps of size $14\times14$ are transformed into $16$ capsules each contains $14\times14$ vectors of $16$ dimensions. Routing from this layer to the ASL Capsule (ASLCap) layer is computed as follows.
\begin{equation}
v_j = \frac{||s_j||^2}{1+||s_j||^2}\frac{s_j}{||s_j||}, s_j = \sum_i{c_{ij}\hat{u}_{j|i}}
\end{equation}
\begin{equation}
c_{ij} = \frac{exp(b_{ij})}{\sum_k{exp(b_{ik})}}, \hat{u}_{j|i}=W_{ij}u_i
\end{equation}
\begin{equation}
b_{ij} \xleftarrow{} b_{ij} + \hat{u}_{j|i}{v_j}
\end{equation}
where $v_j$ represents the vector output of capsule $j$ in ASLCap and $s_j$ is its total input produced by a weighted sum of all predictions from layer below. Next, the signal is passed through a squash non-linearity so that the value is in the range of $[0,1]$. The length of this vector suggests the probability an entity (represented by the capsule) being detected. For example, if the vector $v_j$ is set to represent the sign ``L" then the length of this vector indicates if the sign is actually presented. The output of vector $u_i$ is transformed to the vector $\hat{u}_{j|i}$ by multiplying with a weight matrix $W_{ij}$. The routing coefficient $b_{ij}$ will be increased or decreased based on whether the output $v_j$ has a similar direction with its prediction $\hat{u}_{j|i}$.

%
\subsection{Deep Learning Models}
\label{sec:transferlearning}

This section deals with the integration of Transfer Learning models in Deep Learning architecture for ASL recognition to classify alphabet signs in continuous frames. 
\begin{figure}[!t]
\begin{center}
\includegraphics[width=0.95\textwidth]{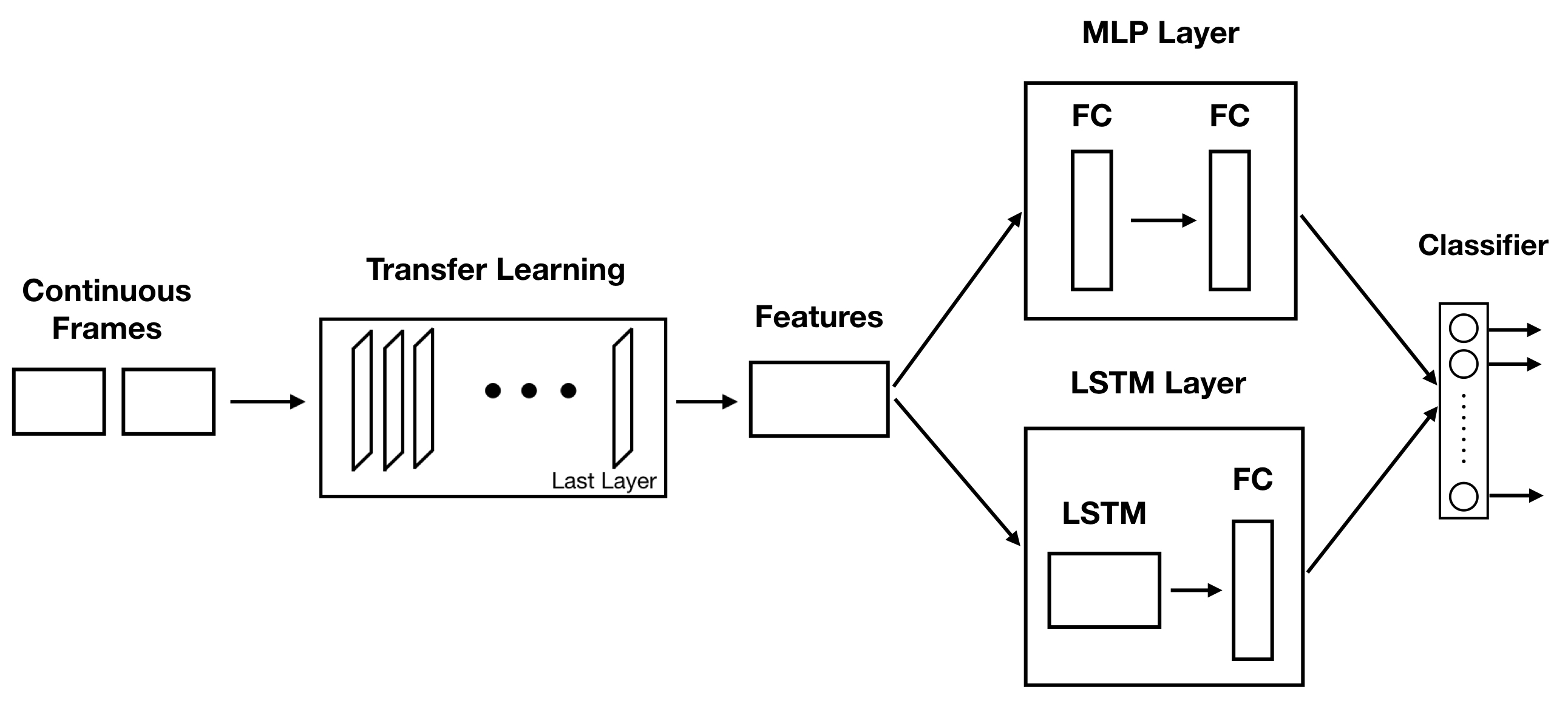}
\caption{Transfer Learning Architecture for ASL}
\label{fig:asl_transferlearning}
\end{center}
\end{figure}
Convolution Neural Networks (ConvNets) were introduced $20$ years ago with the notable architecture from LeNet \cite{lecun1998gradient}. Throughout the time, ConvNets tended to go deeper e.g. VGG ($19$ layers). With recent advances in computer hardware, very deep architectures such as Highway Networks \cite{srivastava2015training} and Residual Networks \cite{he2016deep} have exceeded $100$ layers. Training these models can take days or even months; this gives rise to TL where models are pre-trained on a dataset and re-used on others. In Computer Vision, earlier convolution layers are considered to behave similarly to edge and curve filters. Thus, these frozen weights can be used on our ASL dataset. A modification in the last layer is necessary as the number of signs is $29$ compared to $1000$ categories in ImageNet.

In our research, we include Inception V3, DenseNet V201, NASNet, MobileNet V1 and MobileNet V2 models in DL Architectures. Inception V1 (or often called GoogLeNet to honor LeCun's Networks) was a winner for image classification challenge in ILSVRC 2014 (ImageNet Large Scale Visual Recognition Competition) and achieved top-5 error of $6.67\%$~\cite{szegedy2015going}. The results for Inception V3~\cite{szegedy2016rethinking} and DenseNet V201~\cite{huang2017densely} were $5.6\%$ and $6.34\%/5.54\%$ (on single-crop/10-crop) respectively. 

NASNet was recently introduced and obtained state-of-the-art results on several datasets including Cifar-10, ImageNet and COCO \cite{zoph2018learning}. NASNet comes with two versions one for computers (NASNetLarge) and the other for small devices (NASNetMobile). Since the computation for NASNetLarge takes twice the required time for the slowest models DenseNet V201, we select only NASNetMobile in our experiments. Despite of being made for small devices, the model accomplishes $8.4\%$ top-5 error on ImageNet. Additionally, we choose MobileNets\cite{howard2017mobilenets,sandler2018mobilenetv2} for comparisons within mobile platform.

As shown in Figure \ref{fig:asl_transferlearning}, ASL's signs are illustrated in a video and extracted as a sequence of frames (Please see our demonstration for more information). At the runtime, an active model with trained weights is loaded. Based on each Transfer Learning model, the extracted features have length variations i.e. $2048$, $1920$, $1056$, $1024$ and $1280$ for Inception V3, DenseNet V201, NASNetMobile, MobileNet V1 and V2.

After this step, the flow goes through either MLP Layer or LSTM Layers. The former comprises  two Fully Connected Neural Networks (FC) whereas the latter contains one Long Short Term Memory (LSTM) and one FC. The LSTM has $2048$ units with history's lookback of one for classifying one frame per time. Besides, all FCs are composed by $512$ neurons. We use MLP in our DL models as a baseline for comparison with Vector CNs. This is also interesting to see performance of DL built on other techniques like LSTM.
\section{Experiments and Results}
\label{sec:experiments}
In this section, we first discuss the performances of Vector CNs and DL models on variations of input image size and number of samples. Later, we take two models in DL and compare with two models in Capsule.
\subsection{Experiment 1: Effective of Dataset Size and Image Size on Capsule Accuracy}
To perform this experiment, we scale the Input images to $64 \times 64$, $32 \times 32$ and $16 \times 16$ pixels and use datasets of $\frac{1}{16}$, $\frac{1}{8}$, $\frac{1}{4}$, $\frac{1}{2}$, and a full size. We vary the size of the dataset so that we can observe the effects when the number of samples is small. We exclude dataset of $\frac{1}{32}$ since this yields fewer than $100$ samples per class which may not be enough for recognition. The dataset is split into train and test with the ratio of $70$/$30$. We also use two FCs for Reconstruction, one FC has a half of Input image size and the other has an equal size e.g. when the Input image is $64 \times 64$, the two FCs have sizes of $1024$ ($32 \times 32$) and $4096$ ($64 \times 64$), accordingly.
\begin{figure*}[h]
  \centering
  \subfloat[$\frac{1}{16}$ Dataset]{\includegraphics[width=0.49\textwidth]{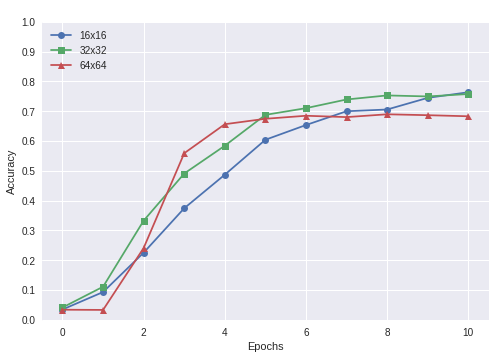}
  \label{fig:capsule_imagedatasize_1_16}}
  \subfloat[All Dataset]{\includegraphics[width=0.49\textwidth]{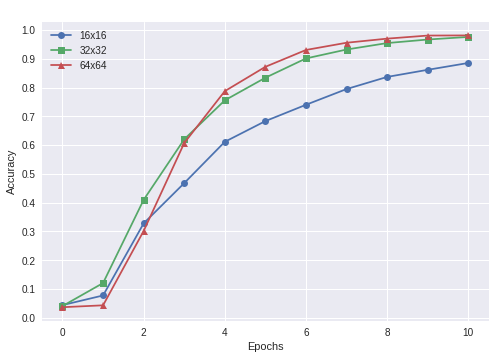}
  \label{fig:capsule_imagedatasize_1_1}}
  \caption{Comparisons of Capsule Networks on Image Sizes}
  \label{fig:capsulenetworks_sizes}
\end{figure*}

Figure \ref{fig:capsule_imagedatasize_1_16} shows a result of this experiment using $\frac{1}{16}$ dataset. We notice that the accuracy for $64 \times 64$ resolution images are lower than for $32 \times 32$ and $16 \times 16$. This contradicts our expectation since with a larger resolution, the image is clearer and should be recognized better. From Figure \ref{fig:capsule_imagedatasize_1_1}, we see a different approach where $64 \times 64$ resolution images perform better than the others. We exclude results for remaining sets to save space.

In summary, these experiments show that a larger image's resolution does not always yield a better accuracy because a small number of samples may affect this performance. With a larger dataset, a higher resolution generally results in a better accuracy. In addition, we can observe that both $64 \times 64$ and $32 \times 32$ image resolutions achieve an accuracy of approximately $0.99$ after $10$ epochs using full dataset.

\subsection{Experiment 2: Comparisons of Deep Learning Models}
\begin{figure*}[h]
  \centering
  \subfloat[All Dataset]{\includegraphics[width=0.49\textwidth]{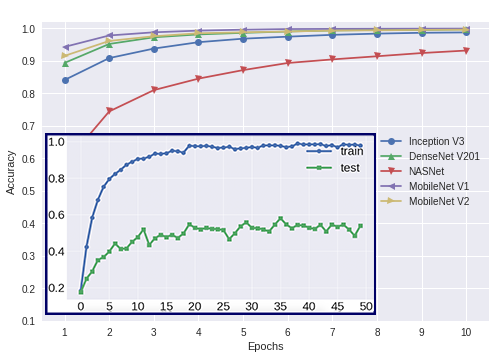}
  \label{fig:MLP_1_1}}
  \subfloat[$\frac{1}{16}$ Dataset]{\includegraphics[width=0.49\textwidth]{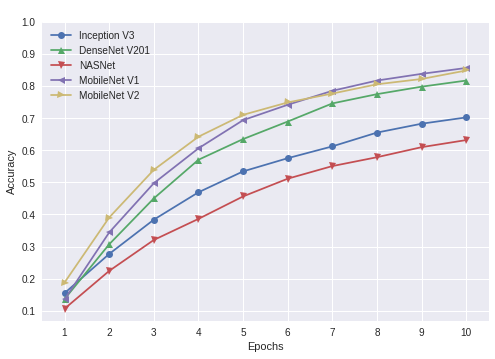}
  \label{fig:MLP_1_5}}

  \caption{Comparisons of Deep Learning Models on MLP. The Figure on the left hand side also includes a result of training and testing on a typical ConvNet (two Convolution layers each with 32 filters, one FC of 128 neurons, Adam optimizer and 50 epochs)}
  \label{fig:transfer_learning_mlp_size_dataset}
\end{figure*}

In this experiment, we use the same variations of dataset size as in Vector CN models and compare all DL models on MLP and LSTM. Figure \ref{fig:MLP_1_1} shows accuracy of all models with MLP and results from a typical ConvNet using full dataset. We can observe that, despite of being simple, the accuracy approaches $100\%$ on training set. However, this yields only near $60\%$ on test. 
\begin{figure*}[!b]
  \centering
  \subfloat[All Dataset]{\includegraphics[width=0.49\textwidth]{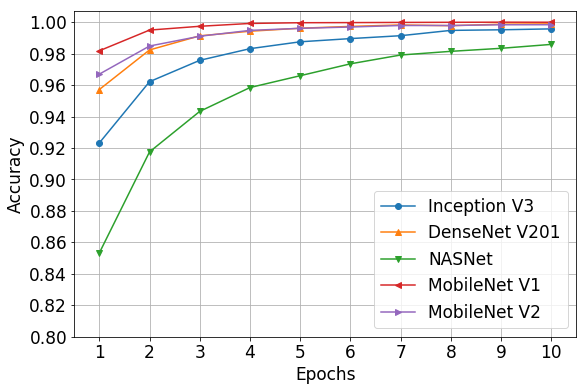}
  \label{fig:LSTM_2_1}}
  \subfloat[1/16 Dataset]{\includegraphics[width=0.49\textwidth]{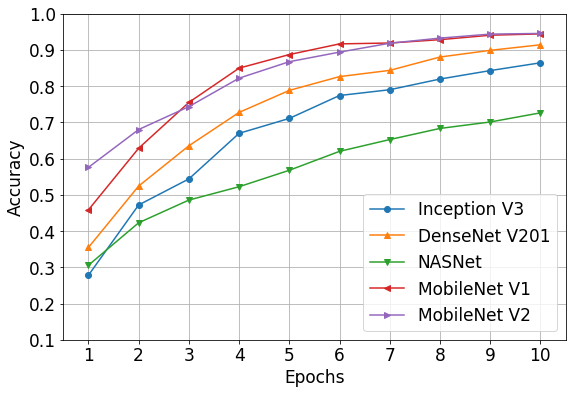}
  \label{fig:LSTM_2_5}}

  \caption{Comparisons of Deep Learning Models on LSTM}
  \label{fig:transfer_learning_lstm_size_dataset}
\end{figure*}

Much to our surprise, two versions of MobileNets both outperform Inception V3 and DenseNet 201. This maybe caused by the efficiency of Deepwise Separable Convolution layers. It also can be seen that DenseNet V201 performs better than Inception V3. Perhaps, the model's blocks in which one layer is connected to all other layers helps to reserve more important information. Additionally, a similar trend can be seen using $\frac{1}{16}$ dataset with accuracy be offset approximately $10\%-20\%$. Moreover, we can observe analogous results using LSTM in Figure \ref{fig:transfer_learning_lstm_size_dataset}. Overall, DL models classified with LSTM show faster convergences.

\subsection{Experiment 3: Comparisons of Vector Capsule Networks and Deep Learning Models on Speed}
In this experiment, we perform speed tests of Vector CNs and DL models on running times for Feature Extraction and Prediction. Regarding the first metric, all models are executed on GPUs of Tesla K80 and P100. Capsules are excluded since extraction of features is not essential.

We observed that with only one ConvNet, the model performed very well on images $64\times64$ but poorly on the target video. This maybe caused by the model learns very well on the training data but distinct backgrounds between the two hamper its performance. For this reason, we add one more Convolution layer without Pooling after Input image to filter better the backgrounds. We did not add more ConvNets since we experienced the vanishing gradient problem with accuracy becoming zeros even after a long training. 

The structure of Vector CNs are changed as follows. For Capsule with Input image size $32 \times 32$ (aka Capsule 32 V1), we use one Convolution layer after the Input layer and four FCs in Reconstruction with sizes of $4 \times 4$, $8 \times 8$, $16 \times 16$ and $32 \times 32$. For Capsule with Input image sizes of $64 \times 64$, we add two Convolution layers where the first layer is followed by a Pooling with size $2 \times 2$ and the second Convolution layer is without a Pooling. We call this is Capsule 32 V2. In reconstruction, we design four FCs with sizes $8 \times 8$, $16 \times 16$, $32 \times 32$ and $64 \times 64$.
\begin{table}[!htb]
\centering
\caption {\label{tab:capsule_transferlearning_speed}Comparisons of Vector CNs and DL Models on Speed}
\begin{tabular}{cl|c|c|c|}
\cline{3-5}
\multicolumn{1}{l}{}                                                                               &                 & \multicolumn{3}{c|}{\textbf{Metrics}}                                                                                                            \\ \cline{2-5} 
\multicolumn{1}{l|}{}                                                                              & \textbf{Models}& \textbf{\begin{tabular}[c]{@{}c@{}}Feature\\Extraction\\(Tesla K80)\end{tabular}} & \textbf{\begin{tabular}[c]{@{}c@{}}Feature\\ Extraction\\ (Tesla P100)\end{tabular}} & \textbf{\begin{tabular}[c]{@{}c@{}}Prediction\\ (CPU 2.7GHz)\end{tabular}} \\ \hline
\multicolumn{1}{|c|}{\multirow{7}{*}{\textbf{\begin{tabular}[c]{@{}c@{}}\rotatebox{90}{Time (s)}\end{tabular}}}}&Inception V3    &3453                                                                                &1910                                                                                 & 0.248                                                               \\ \cline{2-5} 
\multicolumn{1}{|c|}{}                                                                             &DenseNet V201   &6250                                                                                &3199                                                                                 & -                                                                   \\ \cline{2-5} 
\multicolumn{1}{|c|}{}                                                                             &NASNet          &3018                                                                                &2592                                                                                 & -                                                                   \\ \cline{2-5} 
\multicolumn{1}{|c|}{}                                                                             &MobileNet V1    &870                                                                                 &507                                                                                  & 0.070                                                               \\ \cline{2-5} 
\multicolumn{1}{|c|}{}                                                                             &MobileNet V2    &1175                                                                                &758                                                                                  & -                                                                   \\ \cline{2-5} 
\multicolumn{1}{|c|}{}                                                                             &Capsule 32 V1      &-                                                                                   &-                                                                                    & 0.069                                                               \\ \cline{2-5} 
\multicolumn{1}{|c|}{}                                                                             &Capsule 32 V2      &-                                                                                   &-                                                                                    & 0.086                                                               \\ \hline
\end{tabular}
\end{table}

It can be gleaned from Table \ref{tab:capsule_transferlearning_speed} that MobileNet V1 runs faster than any other models on both GPUs whereas DenseNet V201 is the slowest. This clearly shows the advantage of Depthwise Separable Convolution in reducing computation. As we expected, all models for mobiles perform faster than models for computers with an exception that NASNet is slower than Inception V3 on Tesla P100.

On prediction speed comparison, we pickup two models from DL. We select MobileNet V1 because of its best accuracy and fastest speed. Besides, Inception V3 is chosen instead of DenseNet V201 based on a faster speed. We perform this experiment on a computer with 4-CPUs of 2.7GHz. It is unexpected that Capsule 32 V1 is only slightly faster than MobileNet V1 since the model has only $2$ CN layers (and 3 ConvNets) compared to $28$ ConvNets of its counterpart. It is also noticed that Capsule V2 runs three times faster than Inception 32 V3 but quite slower than MobileNet V1.

\subsection{Experiment 4: Comparison of Vector Capsule Networks and Deep Learning Models on Accuracy}
\begin{figure}[h]
  \centering
  \includegraphics[width=0.95\textwidth]{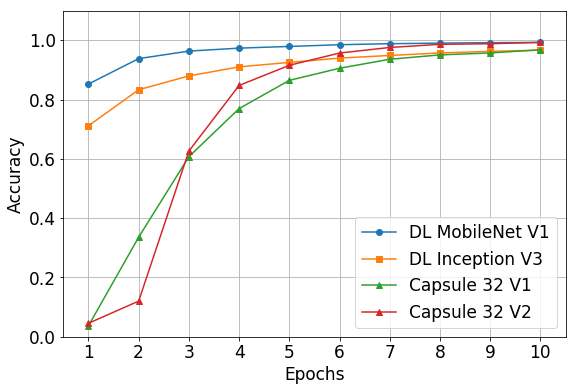}
  \caption{Capsule Networks vs Deep Learning Models on Accuracy}
  \label{fig:capsulevstransferlearning}
\end{figure}

This experiment is designed to compare accuracy of Capsules and Deep Learning architectures. We select MobileNet V1, Inception V3, Capsule 32 V1 and Capsule 32 V2 as discussed in the previous section.

As we can see from Figure \ref{fig:capsulevstransferlearning}, MobileNet V1 and Inception V3 perform better than Capsules in first few epochs. However, as number of epochs increases, Capsule 32 V1 achieves a similar accuracy of Inception V3 and Capsule 32 V2 approaches the accuracy of MobileNet V1.

\section{Conclusions}
\label{sec:conclude}
In this research, we propose to improve the design of Vector CN by extending ConvNet layers after Input image and vary the number of FCs in Reconstruction to accomplish better accuracy. Having a larger image size is critical in our approach since certain signs are similar. A small change in the position of fingers can yield distinct signs. In addition, variation of background contexts may hamper Capsules' performance. Using our approach we attain the goals as follows. First, our results show that Capsule 32 V2 performs comparatively to DL MobileNet V1 on accuracy and Capsule 32 V1 runs a slightly faster than its counterpart. Second, Vector CNs are greater than DL models in the sense that the latter requires pre-trained weights which can take days or even months for training whereas the former can be trained on-the-fly within an hour. Third, as a result of our exploration, we found that MobileNet V1 even though was mainly built for small devices but is superior to all other DL models both on accuracy and speed in this dataset.

Furthermore, we made a demonstration to illustrate our approach where we compare Vector CN V1 with DL MobileNet V1 in an ASL video. The recorded file can be accessed via the link\footnote{http://bit.ly/2O4sJSU}. Although, Vector CN can recognize most of all signs (excluding signs that we mentioned earlier), the model is more sensitive to changes than the DL model. This suggests pooling in the DL may better than just rescaling as in the Vector CN. In addition, the DL performs better although the accuracy (when testing on the ASL dataset) is similar to that of Vector CN. This can indicate that the DL model is more generative. For this reason, additional samples are needed to improve the overall performance.

Although we demonstrate this approach in the context of ASL alphabet signs, the approach has broader applications to any video recognition tasks where each individual frame's information are crucial.

In the future, we plan to redesign CNs' structure to perform on larger images. We also plan to build Vector CNs and DL MobileNet V1 on mobile devices as the accuracy and speed allow these networks to run in realtime.

\section{Acknowledgements}
The reviewers are gratefully acknowledged for their insightful comments. We also thank CISUC - Center of Informatics and Systems of the University of Coimbra for financial support.

\bibliographystyle{splncs04}
\bibliography{ICANN}


\end{document}